\documentclass{article}

\usepackage{microtype}
\usepackage{graphicx}
\usepackage{subcaption}
\usepackage{booktabs} 

\usepackage{hyperref}

\usepackage[absolute,overlay]{textpos}
\usepackage{float}

\usepackage{fancyhdr}

\usepackage{balance}

\usepackage[preprint]{icml2026}

\usepackage{amsmath}
\usepackage{amssymb}
\usepackage{mathtools}
\usepackage{amsthm}
\usepackage{bm}

\usepackage[capitalize,noabbrev]{cleveref}

\usepackage{xcolor}

\usepackage{xurl}

\usepackage{titlesec}

\titleformat{\section}
  {\normalfont\Large\bfseries}  
  {\thesection}{1em}{}

\titleformat{\subsection}
  {\normalfont\large\bfseries}
  {\thesubsection}{1em}{}

\theoremstyle{plain}

\theoremstyle{definition}

\theoremstyle{remark}

\usepackage[textsize=small]{todonotes}

\setuptodonotes{
    color=yellow!30,
    bordercolor=black,
    textcolor=black
}

\begin{document}

\pagestyle{plain}

\thispagestyle{fancy}
\fancyhf{}  
\fancyfoot[L]{\small Morris, M.R. (2026). VET: A Framework for Analyzing AI Discourse. In Proceedings of the Paris
Institute for Advanced Study (Vol. 21). \url{https://doi.org/10.5281/zenodo.20441584} -- \url{https://paris.pias.science/vet-a-framework-for-analyzing-ai-discourse/paris_ias_morris_vet_framework.pdf} -- ISSN 2826-2832/© 2026 Morris, M.R. -- This is an open access article published under the Creative Commons Attribution-NonCommercial 4.0 International Public License (CC BY-NC 4.0)}
\renewcommand{\headrulewidth}{0pt}  

\begin{textblock*}{3cm}(1cm, 1cm)  
  \includegraphics[width=3cm]{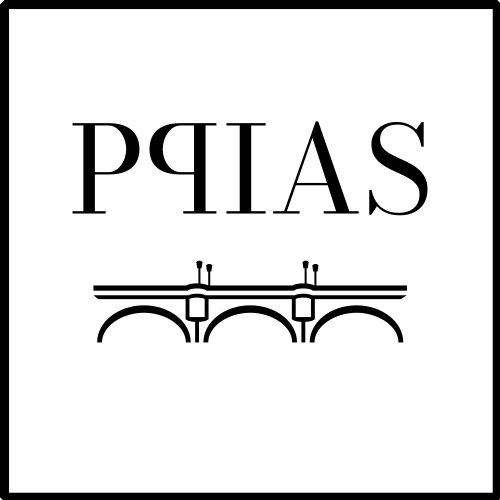}
\end{textblock*}

\onecolumn

\fontsize{12}{14}\selectfont

\vspace*{0.6in}


 \begin{center}
   {\LARGE \textbf{VET: A Framework for Analyzing AI Discourse}} \\ 
   \vspace{0.1in}
   {\bfseries Meredith Ringel Morris} \\
   \vspace{0.1in}
   {The Paul G. Allen School of Computer Science \& Engineering and The Information School \\ University of Washington \\ Seattle, WA, USA} \\
   \vspace{0.1in}
   \texttt{merrie@cs.washington.edu}
 \end{center}

 \vskip 0.3in

\begingroup
 \renewcommand{\footnotetext}[1]{}
 \printAffiliationsAndNotice{}
 \endgroup

\begin{abstract}
 Public discourse on AI has become polarized; exaggerated positions on AI in traditional and social media threaten the development of AI Literacy among the general public. In this article, I introduce the VET Framework, a method for categorizing AI discourse along the dimensions of valence, effectiveness, and trajectory. I show how this framework can be used to identify, compare, and critique prevalent narratives of AI Hype, AI Doom, AI Denial, and AI Normalcy. Using VET, I analyze how each of these four stances exaggerates some aspects of the current state and/or likely evolution of AI, and illustrate how the VET framework can serve as an AI Literacy tool by supporting the ``vetting'' of polarized AI discourse.
\end{abstract}

\section{Introduction}

AI Literacy \cite{unescoLiteracy, literacy2, whatIsLiteracy, AIMagLiteracy} refers to age- and role-appropriate understanding of what modern AI is (\textit{i.e.}, what products and processes use AI), how AI works (\textit{i.e.}, what AI is capable of and what its limitations are), and how AI is changing (\textit{i.e.}, what is the likely trajectory for AI adoption and how will it impact society). As such, AI Literacy is a critical skill for the general public and for policymakers, in order to support appropriate and effective individual, community, corporate, and governmental uses of AI and preparation for societal changes that may result from widespread adoption of emerging AI technologies.  

Formal AI Literacy curricula may play a role in educating target populations (\textit{e.g.}, school-age students, policymakers, workers in job up-skilling or re-training courses). However, traditional media and social media discourse play a key role in shaping knowledge of AI among the general adult population, the large majority of whom do not currently have access to formal AI Literacy content \cite{OECD2025, LAUPICHLER2022100101}.

At the time of writing (early 2026), both traditional news media and social media in the United States present polarized caricatures of both the current state of AI and its future evolution \cite{gilardi2024narratives}; such depictions either exaggerate or downplay the capabilities, benefits, risks, and likely development and adoption trajectory of this emerging class of technologies. While some of the narratives I discuss are present beyond the U.S., AI discourse has high global variance. For example, attitudes toward AI in China and in parts of the Global South are quite different than in the U.S. \cite{maslej2025aiindex}. Attitudes in Europe also differ; for example, Europe's interest in AI sovereignty (having models and data centers that are owned by and located within EU countries) \cite{EUsovereign} is quite different than current anti-datacenter backlash in the United States \cite{pew2026datacenters}. 
Various factors contribute to these regional differences, including government messaging, regulatory frameworks, economic conditions, and political circumstances.

Exaggerated depictions of AI are an impediment to AI Literacy. I propose the VET framework for understanding the current nature of prevalent depictions of AI along the dimensions of \textit{valence}, \textit{effectiveness}, and \textit{trajectory}. Using this framework, I show how different points in this taxonomy represent four common AI media narratives in the U.S.: Hype, Doom, Denial, and Normalcy. Using the VET Framework I compare and contrast these narratives and analyze their strengths and weaknesses, illustrating the framework's utility as an AI Literacy tool to ``vet'' the plausibility of opinions about AI.

\section{VET: A Taxonomy for AI Discourse}

To better understand current media discourse around AI in the United States, I introduce that VET Framework, which considers three facets of beliefs about AI: valence, effectiveness, and trajectory.

\subsection{Valence}

The first dimension helpful for analyzing discourse on AI is the \textit{valence} in attitude toward AI that an author, media organization, social media post, and/or journalistic artifact exhibits. This axis ranges from ascribing extremely negative properties to AI to extremely positive ones. 

Valence can relate to perceived or anticipated impacts of AI. Negative valence is often associated with concerns about impacts such as environmental externalities \cite{li2025thirsty, iea2025electricity}, job loss \cite{briggs2023ai}, impact on human relationships \cite{fang2025aihumanbehaviorsshape, phang2025investigatingaffectiveuseemotional}, the potential for differential impacts on marginalized populations \cite{gallegos2024bias}, or deskilling \cite{kosmyna2025brainchatgptaccumulationcognitive, gerlich2025ai}.  Positive valence tends to be associated with excitement about impacts such as elimination of tedious work \cite{noy2023experimental}, skill expansion through human-AI teaming \cite{brynjolfsson2025generative}, augmented abilities for people with disabilities \cite{lampost, paglialunga2025effectiveness}, novel social support opportunities \cite{genghosts, defreitas2025ai}, and acceleration of scientific and medical discovery \cite{jumper2021alphafold, bai2023ai}. 

Valence can also relate to meta-issues associated with AI, such as attitudes toward technology, privacy, capitalism and wealth inequality, and other political attitudes (\textit{e.g.}, NIMBY-ism, Trumpism \cite{axios2026harrispoll}, pro- or anti-military attitudes, perceptions of China, etc.). 

Finally, valence can relate to positive or negative motivations that some may ascribe to the AI itself, \textit{i.e.}, the (controversial) belief that AI may currently or in the future be sentient \cite{superintelligence, everyoneDies} and may inherently have positive or negative desires as regards human society (\textit{i.e.}, AI may or may not be aligned with human values \cite{alignmentProblem}).     

\subsection{Effectiveness}

A second useful dimension to consider is the attitude toward the \textit{effectiveness} of AI. This dimension ranges from views that consider AI to be a weak and ineffective technology to views that consider AI to be extremely powerful and capable. Effectiveness encompasses considerations of AI's accuracy (\textit{i.e.}, the rate and severity of errors) and reliability (\textit{i.e.}, whether AI performs consistently and has calibration for refusing tasks it cannot perform well) \cite{rabanser2026scienceaiagentreliability}. Note that determinations around accuracy and reliability, while sometimes presented in absolute terms, may be more actionable when contextualized in relative terms, \textit{i.e.}, with respect to a particular baseline, often typical or expert human performance for the same task \cite{merrieJaggedness}. 

Effectiveness can also reflect beliefs about direct comparisons of AI capabilities to human ones, with low effectiveness ratings reflecting the view that AI is not nearly as ``intelligent'' as humans \cite{parrots} and higher effectiveness ratings reflecting the view that AI is approaching milestones such as AGI (Artificial General Intelligence) \cite{morris2025levelsagioperationalizingprogress}.

\subsection{Trajectory}

Finally, the \textit{trajectory} dimension reflects attitudes about the rate of AI progress. The trajectory dimension includes attitudes about the \textit{technical trajectory} of AI -- beliefs about whether technical progress toward milestones such as AGI, robotics with human-level dexterity, level five self-driving vehicles, etc., will be fast or slow. The difference between a predicted ``slow'' versus ``fast'' technical trajectory might be quite large, since this may mean a difference in view on whether the trajectory is likely to be linear or exponential \cite{kaplan2020scalinglawsneurallanguage, lu2026automation}. 

Additionally, this dimension reflects beliefs about the \textit{adoption trajectory} of AI, such as whether individuals, corporations, and governments will incorporate AI into their lives and workflows quickly, or will be delayed by learning curves, regulation, social resistance, or other factors \cite{NormalTech, acemoglu2025macroeconomics, brynjolfsson2021jcurve}. 

Finally, this dimension also reflects beliefs about the \textit{impact trajectory} of AI. The slow end of the impact trajectory scale represents concerns such as Amdahl's Law \cite{amdahl1967validity}, which suggests certain aspects of tasks and processes may act as bottlenecks that limit the speed with which AI can have impact (\textit{e.g.}, while AI may speed up parts of the drug discovery process such as identifying candidate pharmaceuticals, other aspects such as patient trials and the FDA approval process may remain slow and limit the effective speed of end-to-end AI-powered drug discovery pipelines \cite{drugDiscovery}).  The fast end of the impact trajectory scale predicts near-term, pervasive impacts of AI on many aspects of life \cite{darioHype}.

\section{Identifying and Analyzing Prevalent AI Narratives with VET}

\begin{figure*}[t]
  \centering
  \includegraphics[width=\textwidth]{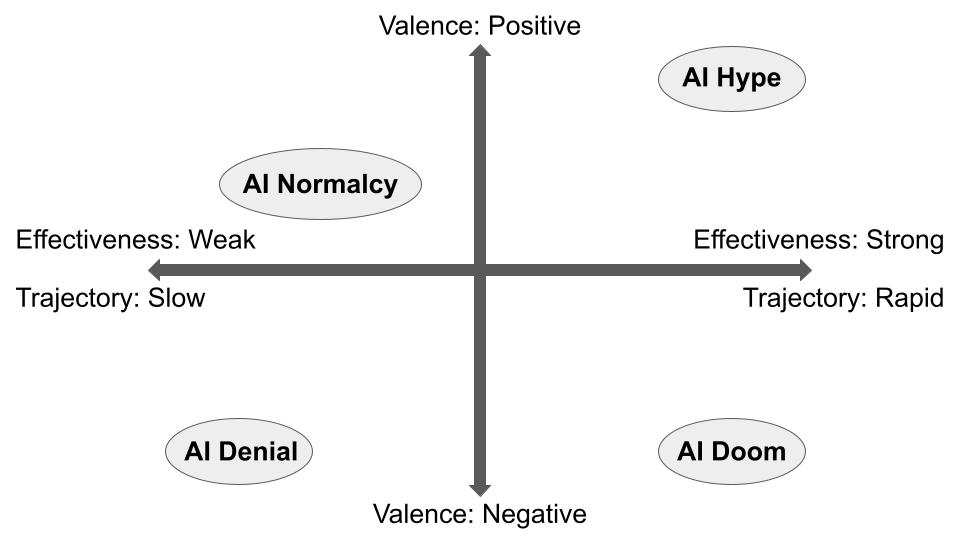} 
  \caption{The VET Framework: An illustration of how the proposed analytic framework of valence, effectiveness, and trajectory helps to identify and compare four prevalent narratives in U.S. discourse about AI. Note that, per the framework, AI Normalcy is less strongly polarized with respect to valence and effectiveness than the other three narratives. Because effectiveness and trajectory are correlated in many narratives (with Normalcy as a counterexample), for simplicity the framework is visualized as a 2D grid.}
  \label{fig:framework}
\end{figure*}

Using the VET Framework, I show how different combinations of of valence, effectiveness, and trajectory identify common narratives about AI, and use the framework to compare and contrast these narratives. Attitudes about AI effectiveness and AI trajectory are correlated for many narratives (with Normalcy as a noteworthy exception that illustrates the value of considering perspectives on effectiveness and trajectory separately), so for simplicity \autoref{fig:framework} depicts the space of AI attitudes as a two by two grid, with valence on the y-axis and a combined effectiveness/trajectory dimension on the x-axis. The four segments of this grid correspond to the four prevalent media attitudes in current U.S. discourse: AI Hype, AI Doom, AI Denial, and AI Normalcy. 



\subsection{AI Hype}

\textit{AI Hype} narratives combine a positive valence for AI with predictions of strong effectiveness and a rapid trajectory of progress, adoption, and impact. Media perspectives aligned with AI Hype imply that Artificial General Intelligence (AGI), a powerful form of AI that can meet or surpass human performance across nearly all tasks \cite{morris2025levelsagioperationalizingprogress}, will arrive within a short timeframe of a few years. AI Hype predicts that this powerful technology will be quickly deployed to usher in a utopian era of harmony and leisure, in which the power of AGI is used to address all of society's ills, cure all disease, and create economic prosperity such that there is a superabundance of benefits for all the world's citizens \cite{intelligenceAge, kurzweil2005singularity}.

AI Hype narratives are frequently associated with the leaders of AI startups and the venture capitalists who fund them, includng OpenAI CEO Sam Altman's essay ``The Gentle Singularity'' \cite{gentleSingularity}, Anthropic CEO Dario Amodei's blog post ``Machines of Loving Grace'' \cite{darioHype}, venture capitalist Vinod Khosla's essay ``A Roadmap to AI Utopia'' \cite{khosla2024roadmap}, and venture capitalist Marc Andreesen's blog post ``The Techno-Optimist Manifesto'' \cite{andreessen2023techno}. Similarly, technology entrepreneur Elon Musk has made predictions that within about a decade, work and money will be made irrelevant by advances in AI and robotics \cite{rogelberg2026musk}. 

This techno-utopian narrative is reflected in the founding charters of several prominent frontier AI companies -- OpenAI was founded as a non-profit with the goal of creating AGI so that it could distribute its benefits to all of humanity \cite{openai2018charter}, Anthropic was founded as a public benefit corporation to ensure that transformative AI would positively benefit humanity \cite{anthropic2021company}, and DeepMind (now Google DeepMind) was founded with a goal of solving intelligence to benefit humanity \cite{mallaby2026infinity}.

Some variants of the AI Hype narrative assert not only that we are rapidly approaching AGI that will positively transform society, but also that these systems have become so advanced that they are, or may soon be, conscious. Philosopher David Chalmers \cite{long2024takingaiwelfareseriously}, former OpenAI Chief Scientist Ilya Sutskever \cite{sutskever2022conscious}, Google VP Blaise Aguera y Arcas \cite{aguera2022llms}, Nobel Laureate Geoffery Hinton \cite{hinton2025lbc}, and other prominent figures have all indicated they think that AI consciousness is possible or even likely. Anthropic has hired an ethicist to investigate the ``AI welfare'' of its Claude AI model \cite{anthropic2025welfare}, taking actions such as cutting off angry user interactions with Claude in case Claude is an entity who may be emotionally harmed by users' vitriolic inputs; Anthropic has also evolved Claude's Constitution \cite{askell2026constitution}, a document of values and rules that guide Claude's operation, to reflect the attitude of a parent conveying ethical norms to its child \cite{perrigo2026teach}.

AI Hype narratives in the media can have a negative impact on the public's AI Literacy by leading them to overtrust the capabilities of current AI systems and overestimate the rate at which technologies will be improved and adopted. Narratives that hype only positive impacts of AI with utopian visions of superabundance risk preventing the public from offering appropriate critiques of AI and developing systems and/or regulations to mitigate potential negative impacts. 

Narratives that exaggerate the effectiveness of current systems as being akin to (or well beyond) human capabilities risk users' misunderstanding important differences between AI systems and humans, and may result in overreliance \cite{merrieJaggedness}, the formation of parasocial emotional relationships \cite{gabriel2024ethicsadvancedaiassistants}, or even the formation of deity-like beliefs about AI \cite{genghosts}, among other risky behaviors. For example, today's models exhibit jagged spikes and valleys in abilities, with a very different distribution of strengths and weaknesses than human populations \cite{merrieJaggedness}; attributing human-like distributions of skills can mislead users to overgeneralize from strong performance in some areas, resulting in misapplying current AI tools to areas for which they are not yet well-suited. Further, narratives predicting extremely fast trajectories to superintelligence and thus superabundance may encourage risky behaviors such as changes in saving, investing, education, and/or employment strategies \cite{quirozgutierrez2026retirement}.

Hype narratives assume an extremely fast trajectory of AI adoption. However, the trajectory of self-driving cars offers a cautionary counter-narrative. While robo-taxi services like Waymo are beginning to be deployed in select cities, largescale rollout of self-driving vehicles has been much slower than initially predicted even though the technology is available and has safety track records outperforming human drivers \cite{waymo2025safetydata}. Negative public attitudes about self-driving vehicles, including concerns that they steal jobs from human drivers, misunderstandings about their safety records, and expectations that they should be held to a higher standard than human drivers, as well as regulatory and infrastructure challenges, have all slowed the deployment of this technology \cite{aaa2025av, lee2025unions, brookings2024av}; a similar fate may befall modern general AI systems, thus slowing the adoption trajectory. 

In the short to medium term, adoption trajectories are bottlenecked by limits on computation such as chip shortages \cite{sanders2026chips} and the need for data center and power grid infrastructure \cite{iea2025electricity, goldmansachs2024power}. Such trajectories are likely even slower in areas such as the Global South, which often lack the infrastructure to run and utilize modern AI at scale \cite{worldbank2025digital}. National security concerns may also cause governing bodies to limit access to powerful future AI models \cite{hendrycks2025superintelligencestrategyexpertversion}, thus limiting their ability to directly and uniformly benefit all of humanity. Some AI organizations and economists \cite{anthropic2025economicpolicy, openai2026industrialpolicy} have put forth proposals for redistributing the wealth AI is forecast to create via mechanisms such as Universal Basic Income, sovereign wealth funds, and/or robot taxes; however, there is currently no concrete public indication that the U.S. government has any serious near-term plans to implement such policy ideas, suggesting that the potential for a fast and positive financial impact trajectory for the general public is likely slower than Hype narratives predict. 

Further, a key aspect of the impact trajectory assumed in Hype narratives is the assumption that technology can solve humanity's problems; however, many problems do not have simple technological solutions, and instead are social or socio-technical in nature. For example, the invention of new mRNA vaccines for COVID-19 did not ``solve'' the COVID pandemic -- some who wanted the vaccine could not get access to it (including many countries in the Global South), and some people who did have access chose not to take the vaccine. Technology alone does not necessarily solve all complex problems.

The attribution of extreme positive valence to AI in Hype discourse is called into question by the AI safety community. Indeed, the challenge of how to ensure that AI is aligned with human values remains an open area of inquiry \cite{alignmentProblem}. Additionally, for AI to be viewed positively at a global scale would require \textit{pluralistic alignment} \cite{sorensen2024roadmappluralisticalignment}, \textit{i.e.}, finding a way to align the values of AI that recognizes individual and cultural variance, since there is no universal definition of correct values. Successful pluralistic alignment remains aspirational as a goal; it is unclear that a single AI system could ever be viewed as a universal positive by diverse global communities with varied value systems. 

Suggestions that current AI is conscious, present in some Hype narratives, are controversial and generally not widely accepted by scientists and philosophers \cite{perceivedConsciousness}. Questions of whether future systems could have some form of consciousness are more ambiguous and a topic of much academic debate; however, current scientific methods cannot determine machine consciousness -- indeed, the definition of consciousness in people in some situations such as newborns or those in a coma is still an open area of scientific inquiry \cite{beingYou}. Focusing instead on how the \textit{perception} of consciousness in AI systems comes about for end-users, and what impacts such attitudes may have, may be a more tractable area for public discussion, particularly as it relates to developing AI Literacy and supporting safe and responsible use of AI \cite{perceivedConsciousness}.

\subsection{AI Doom}

\textit{AI Doom} narratives emerge from assuming a negative valence for AI combined with strong effectiveness and rapid improvement trajectories. Portrayals of AI Doom in the media focus on existential and catastrophic risks of AI. Existential risks are those that would literally result in the extinction of the human race, and might result from a superintelligent AI system \cite{superintelligence} that is unaligned with human values \cite{alignmentProblem}. Catastrophic risks focus on risks that might harm a large number of people, often through the misuse of powerful AI systems by human actors, such as to design and deploy advanced weapons \cite{hendrycks2025superintelligencestrategyexpertversion, suleyman2023wave, kumar2025quantifyingcbrnriskfrontier} or to commit serious cybercrimes \cite{carlini2026mythos, microsoft2024threats, gtig2026may}.

A key proponent of the AI Doom narrative is Eliezar Yudkowsky, who founded the Machine Intelligence Research Institute (MIRI) and recently co-authored a book called \textit{If Anyone Builds it, Everyone Dies} \cite{everyoneDies}. Yudkowsky's book conveys an extreme narrative of AI Doom, assigning a \textit{p(doom)} value of nearly 100\%. Yudkowsky's book uses a series of thought experiments to convince his readers that the creation of superintelligence would result in the extinction of humanity, essentially arguing that superintelligent computing systems would by definition develop their own goals, have the ability to defeat humans in order to advance those goals, and be so complex that humans would be unable to control them. This approach of using thought experiments to illustrate long-tail but extremely dangerous hypothetical scenarios around AI was popularized by the Oxford philosopher Nick Bostrom, whose influential book, \textit{Superintelligence} \cite{superintelligence}, famously discusses the concept of a ``paperclip maximizer'' -- a misaligned, superintelligent AI that destroys humanity in its quest to harvest all of the planet's resources to create paperclips. 

AI Doom perspectives have been espoused by more mainstream figures in the technology industry and academia, as well. One of the most influential proponents of Doom narratives is Geoffrey Hinton, a University of Toronto computer scientist and retired Google executive whose early research on neural network technology earned him the ACM Turing Award and the Nobel Prize in Physics; Hinton has given interviews in which he has indicated his own \textit{p(doom)} as 10\%, 20\%, or even 50\% \cite{hinton2025cbsnews, mallaby2026infinity}. 

In 2023, many scientists and entrepreneurs signed a statement asserting that, ``Mitigating the risk of extinction from AI should be a global priority alongside other societal-scale risks such as pandemics and nuclear war'' \cite{aiSafetyPetition}. Signatories included Microsoft co-founder Bill Gates, Google DeepMind CEO Demis Hassabis, Anthropic CEO Dario Amodei, OpenAI CEO Sam Altman, ACM Turing Award Winner Yoshua Bengio, and many other influential figures. Some signatories may not have intended to advocate fast-trajectory, high-probability AI Doom scenarios, but rather hoped to signal to policymakers that there was a non-zero risk of serious harms from AI, potentially over very long time horizons \cite{mallaby2026infinity}; however, media coverage amplified the negative valence and fast trajectory angle with headlines implying that credible authorities felt AI-caused human extinction may be near (e.g., \cite{roose2023extinction}). 

Media narratives that promulgate AI Doom may lead the public to overestimate the likelihood of extremely low-probability events and to overestimate the capabilities of AI. While considering a broad range of realistic and hypothetical outcomes of AI development is important for AI researchers and policymakers as part of a balanced portfolio of strategic planning and risk management, such narratives may confuse the general public about the near-term probabilities of existential and catastrophic risks.

The extreme negative valence of Doom narratives may result in AI backlash that causes people to overgeneralize about the risks of AI and reject safe and beneficial technologies, such as safe and robust applications of AI to medical, scientific, and cybersecurity advances. In extreme cases, Doom narratives might endanger the safety of those working on AI by inspiring members of the public to take violent action to mitigate hypothesized risks, as in the case of an individual who attacked the home of OpenAI CEO Sam Altman and attempted to set fire to the headquarters of OpenAI in early 2026 \cite{reuters2026altman}. 

By encouraging the public to focus on long-term, long-tail risk scenarios, Doom narratives might negatively impact AI Literacy by usurping focus from more pressing critiques of AI technologies that are likely to be more relevant to members of the general public, such as how AI may impact their physical and mental health, their interpersonal relationships, their education and labor opportunities, the environment, and geopolitics \cite{hanna2023aiharm} -- however, current evidence suggests that while Doom narratives do increase the public's fears about x-risk, they do not seem to lower people's concern regarding other AI risks \cite{hoes2025distraction}. 

Ironically, Doom narratives may also lead people to overtrust AI, because these narratives position AI as being extremely effective. Many Doom narratives also attribute consciousness to current or future AI systems, suggesting they may have their own goals that they choose to pursue at the expense of harming humankind; such narratives may encourage the public to attribute consciousness to AI systems, even when such determinations are not warranted by current scientific consensus \cite{perceivedConsciousness}.

\subsection{AI Denial}

\textit{AI Denial} narratives result from the combination of ascribing a negative valence to AI, low effectiveness, and slow improvement trajectories. AI Denial argues that AI technologies are not and can never be ``intelligent'' in the same way as humans, and therefore cannot work as promised by their developers. Further, Denial narratives ascribe an extremely negative valence to AI, often arguing that the technology will increase societal inequalities, either by increasing income inequality (with the profits from AI going to technology companies or startup founders while others may lose their jobs to low-performing systems that do not work as well as the humans they replaced) or by increasing the gaps between favored and marginalized societal groups (\textit{i.e.}, due to AI fairness concerns such as the use of AI to automate policing, access to social services, hiring processes, and other areas of life in which latent biases in such systems may disadvantage historically marginalized identity groups \cite{gallegos2024bias, ainowinstitute2018}). Note that these arguments appear to be in conflict with each other -- many Denial narratives simultaneously suggest that AI is ineffective, yet nonetheless may be adopted too quickly, thereby causing harm. AI Denial arguments often suggest that the financial, environmental, and/or attentional resources society is directing toward AI development are being wasted on an undesirable and ineffective technology \cite{parrots, crawford2021atlas, narayanan2024snakeoil, thais2024misrepresentedtechnologicalsolutionsimagined}.

An example of this narrative can be found in linguist Emily Bender's and sociologist Alex Hanna's book, \textit{The AI Con} \cite{aiConBook}, which uses semantic arguments to suggest that AI is a ``con'' because it cannot ever have the richness of human intellect and experience, arguing instead that modern AI systems are merely ``stochastic parrots'' \cite{parrots} that use statistical processes to present an illusion of knowledge via fluent text production with no true understanding. Indeed, they argue against the use of the term ``artificial intelligence'' as being intentionally misleading to consumers, and instead they use alternate terminology such as ``synthetic text-extruding machines'' and even ``racist pile of linear algebra'' -- this latter term is meant to reinforce the negative valence they ascribe to AI, \textit{i.e.}, the idea that it is dangerous for society to adopt such systems, which may be ineffective such as by encoding racial biases learned from training data.

Another representative of the AI Denial viewpoint is Gary Marcus, an emeritus professor of psychology and neural technologies at NYU, whose substack \textit{Marcus on AI} \cite{marcus_substack} includes AI Denial commentary. Marcus' posts have provocative titles such as, ``Turns out Generative AI was a scam'' \cite{marcus2026scam} and  ``Why the collapse of the Generative AI bubble may be imminent'' \cite{marcus2024collapse}. Unlike Bender and Hanna, who take issue with the idea that any computational system can possess intelligence, Marcus' form of Denial is more focused -- he believes that the current technical approach underpinning modern generative AI systems will never succeed and therefore assumes weak effectiveness and a slow technical trajectory. Today's systems are primarily driven by taking the transformer architecture \cite{vaswani2023attentionneed} and applying scaling laws \cite{kaplan2020scalinglawsneurallanguage} with large amounts of data and compute, along with reinforcement learning \cite{ouyang2022traininglanguagemodelsfollow, Silver2017}, whereas Marcus argues that neuro-symbolic methods that explicitly encode knowledge and logical reasoning would be a more fruitful technical approach \cite{rebootingAI}.

AI Denial narratives threaten the public's development of AI Literacy by encouraging the public to ignore this important class of emerging technologies because they argue that they have extremely low effectiveness and that the technical trajectory to improvement will also be quite slow. Indeed, some AI Denial commentary actively discourages the public from using technologies such as LLM-powered chatbots \cite{aiConBook}, which may cause those who consume Denial narratives to lack knowledge and skills, thereby growing digital divides. 

While current AI systems are imperfect, the Denial narrative of extremely low effectiveness ignores evidence that AI offers utility for many people and many application areas. ChatGPT was the most quickly adopted technology in history, with the market for LLM-powered chatbot users growing from zero to one hundred million active users within two months \cite{hu2023chatgpt}. AI-powered coding tools have changed the coding industry -- established technology companies and startups alike now strongly encourage or even mandate the use of agentic coding tools because of their proven utility, evidenced by recent systems scoring 87.6\% on SWE-Bench \cite{anthropic2026opus47}, with senior leaders at several companies stating that high percentages of their employees' code is written with or by AI. Modern AI systems have led to scientific advances such as drug discovery \cite{bai2023ai, mullin2026isomorphic} and novel solutions to longstanding challenges in mathematics \cite{alexeev2026primitivesetsvonmangoldt, openAImathdisproof, alon2026remarksdisproofunitdistance}. AI-powered software helps doctors spend more time with patients by performing automated note-taking \cite{tierney2025scribes}, and supports in analyzing medical data and imagery, allowing doctor-AI teams to be more effective at diagnosis than doctors working alone \cite{krakowski2024skincancer, gommers2026masai}. AI is able to identify serious cybersecurity issues in widely available software that human security experts could not find \cite{carlini2026mythos, microsoft2024threats, gtig2026may}. These are just a few examples that illustrate the fallacy of the AI Denial narrative regarding effectiveness of current systems -- while there are many things to improve about AI, to suggest that it does not offer any value is inaccurate. 

AI Denial often suggests that individuals and society should not adopt AI because it does not have perfect accuracy or reliability. For example, AI systems might make mistakes, perhaps even mistakes that are systematically biased against certain demographic groups, due to biases in training data, such as in medical diagnosis or resource allocation \cite{ainowinstitute2018}. However, choosing appropriate baselines for comparison is important to analyzing effectiveness. The threshold for AI deployment for most use cases need not be perfection, but comparison to \textit{current, realistic baselines} for correctness and cost. Human doctors make mistakes in diagnosis; if an AI diagnosis system makes mistakes but makes fewer mistakes than human doctors, this might be of great benefit to many people, even if it is not perfect. Human policymakers exhibit bias (conscious and unconscious) among various demographic groups in resource allocation; even AI systems that have some residual bias may in practice have less biased outcomes than current processes, and such biases may be easier to audit than latent biases in human decision makers \cite{discrimAlg}. Alternatively, society may be willing to accept small tradeoffs in accuracy for some tasks if costs are lowered \cite{li2026economicshumanaicollaboration}. Helping the public understand the baselines for comparison for judging AI's effectiveness, and the dimensions to consider when making decisions about whether to deploy AI (accuracy, costs, baselines, tradeoffs) is a critical component of AI Literacy that is not conveyed through Denial narratives that demand perfection before adoption.

The focus of some AI Denial narratives on the semantics of the term ``Artificial Intelligence'' is perhaps well-intentioned from an AI Literacy standpoint, but is largely moot from a practical standpoint. Artificial Intelligence has become a ``term of art,'' and most academics and companies in the AI space do not make any claims that AI is literally ``intelligent'' in the same way as a human; rather, most focus on the practical capabilities of AI systems rather than amorphous definitions of the semantics of the word ``intelligence.'' However, conveying that AI, while useful for many tasks, need not be conflated with human intelligence or with human-like depictions of AI from science fiction should be an important component of responsible AI discourse.

\subsection{AI Normalcy}

\textit{AI Normalcy} narratives have neutral-to-positive assumptions about AI valence, but combine this with moderate effectiveness and slow improvement, adoption, and impact trajectories. Essentially, AI Normalcy suggests that today's improvements in AI should not be viewed differently than past innovations, that they will not radically disrupt society, and that therefore there is no urgency for individuals, communities, organizations, or governments to take drastic actions regarding AI preparedness.

A representative example of this narrative is Princeton computer scientists' Arvind Narayanan's and Sayash Kapoor's 2025 essay, ``AI as Normal Technology'' \cite{NormalTech}, which argues that while modern AI technologies have made some noteworthy progress, these technologies should be viewed similarly to other past innovations such as electricity or personal computers, which required decades to reach mass adoption and demonstrate concrete productivity gains. MIT Roboticist Rodney Brooks is another prominent AI Normalcy advocate, arguing that short term predictions about technologies' impacts tend to be overblown, whereas long-term adoption and uses (over the course of decades) can be unanticipated \cite{brooks7sins}. 

Economic analyses that anticipate labor replacement due to AI as likely to be relatively slow \cite{brynjolfsson2021jcurve}, that any job losses will be offset by gains from new classes of AI-enabled jobs as has been the pattern with past technologies \cite{autor2015jobs, wef2025jobs}, or that AI-enhanced productivity will drive further demand for labor rather than replace it (i.e., a Jevons Paradox) \cite{jevons1865coal} are also AI Normalcy positions. Invocation of the Solow Paradox (the idea that ``You can see the computer age everywhere but in the productivity statistics'') \cite{triplett1999solow} is another example of an AI Normalcy stance. 

The AI Normalcy narrative thus argues that adoption and impact trajectories are likely to be slow, due to the need for individuals and businesses to understand how to integrate novel AI technologies into their workflows effectively, for the technologies to become reliable and cheap enough that it is cost effective to use them at scale, and due to the likely need to overcome societal frictions to fast adoption such as those introduced by labor unions, regulators, or public backlash. 

Note that AI Normalcy is less ``extreme'' per the VET framework than the Hype, Doom, and Denial narratives. In particular, the Normalcy narrative tends to view effectiveness as moderate rather than weak, and is relatively neutral regarding the valence of AI.  The Normalcy narrative assumes a very slow adoption and impact trajectory, however, as well as a relatively slow technical trajectory -- for instance, expressing skepticism that AGI or superintelligence are meaningfully defined terms and/or realistic technical goals \cite{NormalTech}. 

This nuanced analysis of Normalcy narratives demonstrates how the VET Framework goes beyond prior taxonomies of AI perspectives such as Gilardi et al.'s ontology of existential risk, accelerationist, immediate risk, and balanced risk \cite{gilardi2024narratives} -- AI Normalcy arguments would likely be categorized as ``balanced risk'' in their framework, suggesting such views are the desired and accurate narrative. However, with the VET framework, we see that while AI Normalcy is less extreme in its positioning regarding valence and effectiveness than other prevalent narratives, it nonetheless poses risks to AI Literacy because of its slow positioning along the trajectory dimension. Normalcy narratives risk a public that is unprepared for the rate and degree of personal and societal changes that many experts expect AI to bring \cite{leap2025}. 

People are generally bad at understanding exponential growth curves \cite{wagenaar1975misperception}; the Normalcy narrative projects a non-exponential adoption and impact trajectory of AI, in line with previous technologies. However, current evidence suggests this view of AI's trajectory is inaccurate. For example, the recent case of Anthropic's Mythos model needing to be withheld because it was able to identify a large number of critical cybersecurity flaws \cite{carlini2026mythos} is an example of AI creating a serious and rapid step-change in cyber capabilities, requiring national security responses and potentially having serious geopolitical implications. The recent successes of AI in discovering novel solutions to longstanding open challenges in mathematics \cite{alexeev2026primitivesetsvonmangoldt, openAImathdisproof, alon2026remarksdisproofunitdistance} is another hint that this technology is on a much steeper impact trajectory than Normalcy suggests. 

Indeed, another piece of evidence that Normalcy discourse likely underestimates the technical trajectory of AI is that frontier AI labs are openly pursing a strategy of \textit{recursive self-improvement} \cite{zelikman2024selftaughtoptimizerstoprecursively} for AI research and development, in which they invest in optimizing AI tools for writing code and performing AI research. The goal of recursive self-improvement is to bootstrap the rate of AI development, potentially enabling a \textit{fast takeoff} \cite{superintelligence} or \textit{intelligence explosion} \cite{good1965ultraintelligent} of AI capability since AI that can speed up its own development will logically accelerate future AI advances. This strategy is already bearing fruit, as evidenced by the acceleration of utility of agentic coding tools \cite{carlini2026mythos, anthropic2026opus47}, which have now been widely adopted throughout the AI industry. The impacts of these changes are being reflected not only in the pace of AI development, but in changes to the labor market for software developers, including substantial layoffs at some software companies that are choosing to rely more on AI coding tools \cite{angelo2026block, npr2026meta} and a slowdown in hiring for junior coding positions \cite{brynjolfsson2025canaries}. Such developments provide counter-evidence to the narrative that AI will mature and be adopted at a rate on par with prior technologies, providing early signals of a faster-than-anticipated adoption trajectory in some sectors. 

While bottlenecks to adoption certainly exist, including social issues such as AI backlash among some segments of the general public \cite{pew2025public}, potential future changes in the regulatory landscape, and Amdahl's Law \cite{amdahl1967validity} bottlenecks on the extent to which AI alone can speed up particular industries or processes, the Normalcy narrative risks fomenting complacency among the public and policymakers by leading them to underestimate the time they may have available to prepare for and adapt to likely impacts of AI. This might include both personal preparedness decisions such as re-skilling and collective preparedness decisions such as advocating for particular policies or regulations.

The VET framework also offers insight into why Normalcy narratives may be less prevalent in the media ecosystem than Hype, Doom, or Denial. Traditional and social media tend to amplify more extreme narratives and narratives with negative valence \cite{outgroupAnimosity, moralDiffusion, Robertson2023} -- because Normalcy is less extreme in its predictions about the valence, effectiveness, and trajectory of AI than competing narratives, it may receive less media traction. 

\section{Conclusion}

In this article, I introduced the VET Framework, a new tool for supporting AI Literacy by providing a method of analyzing AI discourse using the dimensions of valence, effectiveness, and trajectory. \textit{Valence} reflects whether an AI narrative depicts AI as negative or positive, including considerations of AI's current and future impacts, meta-issues such as attitudes toward the technology industry, privacy, and capitalism, and even the positive or negative motivations that a hypothesized sentient AI might itself possess. \textit{Effectiveness} encompasses whether an AI narrative positions AI as weak or strong, including considerations of the accuracy and reliability of AI systems, potentially with respect to human baselines. \textit{Trajectory} considers whether a narrative positions AI progress as slow or rapid, including the technical trajectory for AI progress, the adoption trajectory of AI by individuals and organizations, and the rate at which AI will impact society. 

In this article, I have demonstrated how the framework allows for identifying, comparing, and analyzing prevailing media narratives about AI by using the framework to characterize AI Hype, AI Doom, AI Denial, and AI Normalcy based on their depictions of AI's valence, effectiveness, and trajectory. This framework can support AI Literacy by offering a method to systematically ``vet'' the quality of arguments about AI using these three analytical dimensions. I hope that the VET Framework can be a useful tool for policymakers, the media, and the general public to support nuanced discussion and analysis of both the current state of AI and its likely future.

\section*{Acknowledgements}

This article benefited from a fellowship at the Paris Institute for Advanced Study (France), with the financial support of the French State, programme \textit{Investissements d’avenir} managed by the Agence Nationale de la Recherche (ANR-11-LABX-0027-01 Labex RFIEA+).

I would like to thank John Krumm, Siddharth Suri, Iason Gabriel, Michael Terry, Jacy Reese Anthis, Shriya Sekhsaria, Pierre Dillenbourg, and Jan Eißfeldt for helpful conversations and feedback that influenced this work. I would especially like to thank Dan Morris, whose support made my writing residency at Paris IAS possible. 

Note that this work represents the opinion of the author, and is not an official policy statement or position of any institutions with which she is associated.


\balance

\bibliography{example_paper}

@book{mallaby2026infinity,
  author    = {Mallaby, Sebastian},
  title     = {The Infinity Machine: {Demis Hassabis, DeepMind, and the Quest for Superintelligence}},
  publisher = {Penguin Press},
  year      = {2026},
  address   = {New York},
  isbn      = {9780593831847},
  pages     = {480}
}

@misc{roose2023extinction,
  author       = {Roose, Kevin},
  title        = {A.I.\ Poses `Risk of Extinction,' Industry Leaders Warn},
  year         = {2023},
  month        = {May},
  day          = {30},
  howpublished = {\textit{The New York Times}.
                  \url{https://www.nytimes.com/2023/05/30/technology/ai-chatbot-extinction-risk.html}}
}

@article{gilardi2024narratives,
  author    = {Gilardi, Fabrizio and Kasirzadeh, Atoosa and Bernstein, Abraham and Staab, Steffen and Gohdes, Anita},
  title     = {We need to understand the effect of narratives about generative {AI}},
  journal   = {Nature Human Behaviour},
  volume    = {8},
  number    = {12},
  pages     = {2251--2252},
  year      = {2024},
  month     = dec,
  publisher = {Nature Publishing Group},
  doi       = {10.1038/s41562-024-02026-z},
  issn      = {2397-3374},
  note      = {Published online 21 October 2024}
}

@Article{Robertson2023,
author={Robertson, Claire E.
and Pr{\"o}llochs, Nicolas
and Schwarzenegger, Kaoru
and P{\"a}rnamets, Philip
and Van Bavel, Jay J.
and Feuerriegel, Stefan},
title={Negativity drives online news consumption},
journal={Nature Human Behaviour},
year={2023},
month={May},
day={01},
volume={7},
number={5},
pages={812-822},
issn={2397-3374},
doi={10.1038/s41562-023-01538-4},
url={https://doi.org/10.1038/s41562-023-01538-4}
}

@article{
outgroupAnimosity,
author = {Steve Rathje  and Jay J. Van Bavel  and Sander van der Linden },
title = {Out-group animosity drives engagement on social media},
journal = {Proceedings of the National Academy of Sciences},
volume = {118},
number = {26},
pages = {e2024292118},
year = {2021},
doi = {10.1073/pnas.2024292118},
URL = {https://www.pnas.org/doi/abs/10.1073/pnas.2024292118},
eprint = {https://www.pnas.org/doi/pdf/10.1073/pnas.2024292118}
}

@article{
moralDiffusion,
author = {William J. Brady  and Julian A. Wills  and John T. Jost  and Joshua A. Tucker  and Jay J. Van Bavel },
title = {Emotion shapes the diffusion of moralized content in social networks},
journal = {Proceedings of the National Academy of Sciences},
volume = {114},
number = {28},
pages = {7313-7318},
year = {2017},
doi = {10.1073/pnas.1618923114},
URL = {https://www.pnas.org/doi/abs/10.1073/pnas.1618923114},
eprint = {https://www.pnas.org/doi/pdf/10.1073/pnas.1618923114}
}

@techreport{leap2025,
    author = {Murphy, Connacher and Rosenberg, Josh and Canedy, Jordan and Jacobs, Zach and Flechner, Nadja and Britt, Rhiannon and Pan, Alexa and Rogers-Smith, Charlie and Mayland, Dan and Buffington, Cathy and Kučinskas, Simas and Coston, Amanda and Kerner, Hannah and Pierson, Emma and Rabbany, Reihaneh and Salganik, Matthew and Seamans, Robert and Su, Yu and Tramèr, Florian and Hashimoto, Tatsunori and Narayanan, Arvind and Tetlock, Philip E. and Karger, Ezra},
    title = {The Longitudinal Expert AI Panel: Understanding Expert Views on AI Capabilities, Adoption, and Impact},
    institution = {Forecasting Research Institute},
    type = {Working paper},
    number = {5},
    url = {https://leap.forecastingresearch.org/reports/wave3},
    urldate = {2026-05-22},
    year = {2025}
  }

@article{wagenaar1975misperception,
  author  = {Wagenaar, Willem A. and Sagaria, Sabato D.},
  title   = {Misperception of Exponential Growth},
  journal = {Perception \& Psychophysics},
  volume  = {18},
  number  = {6},
  pages   = {416--422},
  year    = {1975},
  doi     = {10.3758/BF03204114}
}

@article{discrimAlg,
    author = {Kleinberg, Jon and Ludwig, Jens and Mullainathan, Sendhil and Sunstein, Cass R},
    title = {Discrimination in the Age of Algorithms},
    journal = {Journal of Legal Analysis},
    volume = {10},
    pages = {113-174},
    year = {2018},
    month = {12},
    issn = {2161-7201},
    doi = {10.1093/jla/laz001},
    url = {https://doi.org/10.1093/jla/laz001},
    eprint = {https://academic.oup.com/jla/article-pdf/doi/10.1093/jla/laz001/30132964/laz001.pdf},
}

@misc{li2026economicshumanaicollaboration,
      title={Economics of Human and AI Collaboration: When is Partial Automation More Attractive than Full Automation?}, 
      author={Wensu Li and Atin Aboutorabi and Harry Lyu and Kaizhi Qian and Martin Fleming and Brian C. Goehring and Neil Thompson},
      year={2026},
      eprint={2603.29121},
      archivePrefix={arXiv},
      primaryClass={econ.GN},
      url={https://arxiv.org/abs/2603.29121}, 
}

@misc{waymo2025safetydata,
  author       = {{Waymo}},
  title        = {Safety Impact Hub},
  year         = {2025},
  howpublished = {\url{https://waymo.com/safety/impact/}}
}

@misc{brookings2024av,
  author       = {West, Darrell M.},
  title        = {The Evolving Safety and Policy Challenges of
                  Self-Driving Cars},
  year         = {2024},
  month        = jul,
  day          = {31},
  howpublished = {Brookings Institution.
                  \url{https://www.brookings.edu/articles/the-evolving-safety-and-policy-challenges-of-self-driving-cars/}}
}

@misc{aaa2025av,
  author       = {{American Automobile Association}},
  title        = {Fear in Self-Driving Vehicles Persists},
  year         = {2025},
  month        = feb,
  day          = {25},
  howpublished = {{AAA} Newsroom.
                  \url{https://newsroom.aaa.com/2025/02/aaa-fear-in-self-driving-vehicles-persists/}}
}

@misc{lee2025unions,
  author       = {Lee, Timothy B.},
  title        = {Unions Want to Ban Driverless Taxis---Will Democratic
                  Leaders Say Yes?},
  year         = {2025},
  month        = aug,
  day          = {7},
  howpublished = {Understanding {AI}.
                  \url{https://www.understandingai.org/p/unions-want-to-ban-driverless-taxiswill}}
}

@techreport{sanders2026chips,
  author      = {Sanders, James and Egan, Janet and Madigan, Rory},
  title       = {American {AI} Companies Can't Get Enough Chips},
  institution = {Center for a New American Security ({CNAS})},
  year        = {2026},
  month       = may,
  day         = {7},
  url         = {https://www.cnas.org/publications/reports/american-ai-companies-cant-get-enough-chips}
}

@misc{goldmansachs2024power,
  author       = {Schneider, James},
  title        = {{AI} to Drive 165\% Increase in Data Center Power
                  Demand by 2030},
  year         = {2024},
  howpublished = {Goldman Sachs Research.
                  \url{https://www.goldmansachs.com/insights/articles/ai-to-drive-165-increase-in-data-center-power-demand-by-2030}}
}

@techreport{worldbank2025digital,
  author      = {{World Bank}},
  title       = {Digital Progress and Trends Report 2025},
  institution = {World Bank},
  year        = {2025},
  month       = nov,
  url         = {https://openknowledge.worldbank.org/entities/publication/8f5d2cb9-92d4-42fd-a4ad-fa538f081488}
}

@techreport{brynjolfsson2025canaries,
  author      = {Brynjolfsson, Erik and Chandar, Bharat and Chen, Ruyu},
  title       = {Canaries in the Coal Mine? Six Facts about the Recent
                 Employment Effects of Artificial Intelligence},
  institution = {Stanford Digital Economy Lab},
  year        = {2025},
  month       = nov,
  url         = {https://digitaleconomy.stanford.edu/publications/canaries-in-the-coal-mine/}
}

@misc{npr2026meta,
  author       = {John Ruwitch},
  title        = {Meta Slashes 8,000 Jobs as It Pivots Towards {AI}},
  year         = {2026},
  month        = may,
  day          = {20},
  howpublished = {\textit{NPR}.
                  \url{https://www.npr.org/2026/05/20/nx-s1-5826917/meta-layoffs-ai-jobs}}
}

@misc{angelo2026block,
  author       = {Angelo, Jake},
  title        = {Block {CEO} Jack Dorsey Lays Off Nearly Half of His
                  Staff Because of {AI} and Predicts Most Companies
                  Will Make Similar Cuts in the Next Year},
  year         = {2026},
  month        = feb,
  day          = {27},
  howpublished = {\textit{Fortune}.
                  \url{https://fortune.com/2026/02/27/block-jack-dorsey-ceo-xyz-stock-square-4000-ai-layoffs/}}
}

@misc{quirozgutierrez2026retirement,
  author       = {Quiroz-Gutierrez, Marco},
  title        = {Elon Musk Says Saving for Retirement Is Irrelevant
                  Because {AI} Is Going to Create a World of Abundance:
                  `It Won't Matter'},
  year         = {2026},
  month        = jan,
  day          = {12},
  howpublished = {\textit{Fortune}.
                  \url{https://fortune.com/2026/01/12/elon-musk-retirement-savings-irrelevant-ai-robots-abundance/}}
}

@misc{anthropic2025economicpolicy,
  author       = {{Anthropic}},
  title        = {Preparing for {AI}'s Economic Impact:
                  Exploring Policy Responses},
  year         = {2025},
  month        = oct,
  day          = {14},
  howpublished = {\url{https://www.anthropic.com/research/economic-policy-responses}}
}

@misc{openai2026industrialpolicy,
  author       = {{OpenAI}},
  title        = {Industrial Policy for the Intelligence Age:
                  Ideas to Keep People First},
  year         = {2026},
  month        = apr,
  day          = {6},
  howpublished = {\url{https://openai.com/index/industrial-policy-for-the-intelligence-age/}}
}

@article{autor2015jobs,
  author  = {Autor, David H.},
  title   = {Why Are There Still So Many Jobs? The History and Future
             of Workplace Automation},
  journal = {Journal of Economic Perspectives},
  volume  = {29},
  number  = {3},
  pages   = {3--30},
  year    = {2015},
  month   = aug,
  doi     = {10.1257/jep.29.3.3}
}

@techreport{wef2025jobs,
  author      = {{World Economic Forum}},
  title       = {Future of Jobs Report 2025},
  institution = {World Economic Forum},
  year        = {2025},
  month       = jan,
  day         = {8},
  url         = {https://reports.weforum.org/docs/WEF_Future_of_Jobs_Report_2025.pdf}
}

@book{jevons1865coal,
  author    = {Jevons, William Stanley},
  title     = {The Coal Question: An Inquiry Concerning the Progress
               of the {Britain} and the Probable Exhaustion of Our
               Coal-Mines},
  year      = {1865},
  publisher = {Macmillan},
  address   = {London}
}

@misc{anthropic2026opus47,
  author       = {{Anthropic}},
  title        = {Introducing {Claude Opus} 4.7},
  year         = {2026},
  month        = apr,
  howpublished = {\url{https://www.anthropic.com/news/claude-opus-4-7}}
}

@misc{mullin2026isomorphic,
  author       = {Mullin, Emily},
  title        = {{AI}-Designed Drugs by a {DeepMind} Spinoff Are Headed
                  to Human Trials},
  year         = {2026},
  month        = apr,
  day          = {24},
  howpublished = {\textit{Wired}.
                  \url{https://www.wired.com/story/wired-health-2026-how-ai-is-powering-drug-discovery-max-jaderberg/}}
}

@article{tierney2025scribes,
  author  = {Tierney, Aaron A. and Gayre, Gregg and Hoberman, Brian
             and Mattern, Britt and Ballesca, Manuel and {Wilson Hannay}, Sarah
             and Castilla, Kate and Lau, Cindy and Kipnis, Patricia
             and Liu, Vincent and Lee, Kristine},
  title   = {Ambient Artificial Intelligence Scribes: Learnings after
             1 Year and over 2.5 Million Uses},
  journal = {NEJM Catalyst},
  year    = {2025},
  doi     = {10.1056/CAT.25.0040}
}

@article{krakowski2024skincancer,
  author  = {Krakowski, Isabelle and Kim, Jiyeong and Cai, Zhuo Ran
             and Daneshjou, Roxana and Lapins, Jan and Eriksson, Hanna
             and Lykou, Anastasia and Linos, Eleni},
  title   = {Human-{AI} Interaction in Skin Cancer Diagnosis:
             A Systematic Review and Meta-Analysis},
  journal = {npj Digital Medicine},
  volume  = {7},
  pages   = {78},
  year    = {2024},
  month   = apr,
  doi     = {10.1038/s41746-024-01031-w}
}

@article{gommers2026masai,
  author  = {Gommers, Jessie and Hernstr{\"o}m, Veronica and Josefsson, Viktoria
             and Sartor, Hanna and Schmidt, David and Hjelmgren, Annie
             and Larsson, Anna-Maria and Hofvind, Solveig and Andersson, Ingvar
             and Rosso, Aldana and Hagberg, Oskar and L{\aa}ng, Kristina},
  title   = {Interval Cancer, Sensitivity, and Specificity Comparing
             {AI}-Supported Mammography Screening with Standard Double
             Reading without {AI} in the {MASAI} Study: A Randomised,
             Controlled, Non-Inferiority, Single-Blinded, Population-Based,
             Screening-Accuracy Trial},
  journal = {The Lancet},
  volume  = {407},
  number  = {10527},
  pages   = {505--514},
  year    = {2026},
  month   = jan,
  doi     = {10.1016/S0140-6736(25)02464-X}
}

@misc{openAImathdisproof,
    url={https://cdn.openai.com/pdf/74c24085-19b0-4534-9c90-465b8e29ad73/unit-distance-proof.pdf},
    author={{OpenAI}},
    month={May},
    year={2026},
    title={Planar Point Sets with Many Unit Distances}
}

@misc{alon2026remarksdisproofunitdistance,
      title={Remarks on the disproof of the unit distance conjecture}, 
      author={Noga Alon and Thomas F. Bloom and W. T. Gowers and Daniel Litt and Will Sawin and Arul Shankar and Jacob Tsimerman and Victor Wang and Melanie Matchett Wood},
      year={2026},
      eprint={2605.20695},
      archivePrefix={arXiv},
      primaryClass={math.CO},
      url={https://arxiv.org/abs/2605.20695}, 
}

@misc{alexeev2026primitivesetsvonmangoldt,
      title={Primitive sets and von {Mangoldt} chains: {Erd\H{o}s} Problem \#1196 and beyond}, 
      author={Boris Alexeev and Kevin Barreto and Yanyang Li and Jared Duker Lichtman and Liam Price and Jibran Iqbal Shah and Quanyu Tang and Terence Tao},
      year={2026},
      eprint={2605.00301},
      archivePrefix={arXiv},
      primaryClass={math.NT},
      url={https://arxiv.org/abs/2605.00301}, 
}

@misc{sorensen2024roadmappluralisticalignment,
      title={A Roadmap to Pluralistic Alignment}, 
      author={Taylor Sorensen and Jared Moore and Jillian Fisher and Mitchell Gordon and Niloofar Mireshghallah and Christopher Michael Rytting and Andre Ye and Liwei Jiang and Ximing Lu and Nouha Dziri and Tim Althoff and Yejin Choi},
      year={2024},
      eprint={2402.05070},
      archivePrefix={arXiv},
      primaryClass={cs.AI},
      url={https://arxiv.org/abs/2402.05070}, 
}

@article{triplett1999solow,
  author    = {Jack E. Triplett},
  title     = {The Solow Productivity Paradox: What Do Computers Do to Productivity?},
  journal   = {Canadian Journal of Economics},
  volume    = {32},
  number    = {2},
  pages     = {309--334},
  year      = {1999},
  month     = apr,
  doi       = {10.2307/136425}
}

@incollection{good1965ultraintelligent,
  author    = {Good, Irving John},
  title     = {Speculations Concerning the First Ultraintelligent Machine},
  booktitle = {Advances in Computers},
  editor    = {Alt, Franz L. and Rubinoff, Morris},
  volume    = {6},
  pages     = {31--88},
  year      = {1965},
  publisher = {Academic Press},
  address   = {New York},
  doi       = {10.1016/S0065-2458(08)60418-0}
}

@misc{zelikman2024selftaughtoptimizerstoprecursively,
      title={Self-Taught Optimizer (STOP): Recursively Self-Improving Code Generation}, 
      author={Eric Zelikman and Eliana Lorch and Lester Mackey and Adam Tauman Kalai},
      year={2024},
      eprint={2310.02304},
      archivePrefix={arXiv},
      primaryClass={cs.CL},
      url={https://arxiv.org/abs/2310.02304}, 
}

@misc{vaswani2023attentionneed,
      title={Attention Is All You Need}, 
      author={Ashish Vaswani and Noam Shazeer and Niki Parmar and Jakob Uszkoreit and Llion Jones and Aidan N. Gomez and Lukasz Kaiser and Illia Polosukhin},
      year={2023},
      eprint={1706.03762},
      archivePrefix={arXiv},
      primaryClass={cs.CL},
      url={https://arxiv.org/abs/1706.03762}, 
}

@misc{ouyang2022traininglanguagemodelsfollow,
      title={Training language models to follow instructions with human feedback}, 
      author={Long Ouyang and Jeff Wu and Xu Jiang and Diogo Almeida and Carroll L. Wainwright and Pamela Mishkin and Chong Zhang and Sandhini Agarwal and Katarina Slama and Alex Ray and John Schulman and Jacob Hilton and Fraser Kelton and Luke Miller and Maddie Simens and Amanda Askell and Peter Welinder and Paul Christiano and Jan Leike and Ryan Lowe},
      year={2022},
      eprint={2203.02155},
      archivePrefix={arXiv},
      primaryClass={cs.CL},
      url={https://arxiv.org/abs/2203.02155}, 
}

@Article{Silver2017,
author={Silver, David
and Schrittwieser, Julian
and Simonyan, Karen
and Antonoglou, Ioannis
and Huang, Aja
and Guez, Arthur
and Hubert, Thomas
and Baker, Lucas
and Lai, Matthew
and Bolton, Adrian
and Chen, Yutian
and Lillicrap, Timothy
and Hui, Fan
and Sifre, Laurent
and van den Driessche, George
and Graepel, Thore
and Hassabis, Demis},
title={Mastering the game of Go without human knowledge},
journal={Nature},
year={2017},
month={Oct},
day={01},
volume={550},
number={7676},
pages={354-359},
issn={1476-4687},
doi={10.1038/nature24270},
url={https://doi.org/10.1038/nature24270}
}

@article{hu2023chatgpt,
  author  = {Krystal Hu},
  title   = {ChatGPT sets record for fastest-growing user base - analyst note},
  journal = {Reuters},
  year    = {2023},
  month   = feb,
  day     = {1},
  url     = {https://www.reuters.com/technology/chatgpt-sets-record-fastest-growing-user-base-analyst-note-2023-02-01/}
}

@article{reuters2026altman,
  author  = {{Reuters}},
  title   = {Man charged after Molotov cocktail attack on OpenAI CEO Sam Altman's home},
  journal = {Reuters},
  year    = {2026},
  month   = apr,
  day     = {13},
  url     = {https://www.reuters.com/legal/government/man-charged-after-molotov-cocktail-attack-openai-ceo-sam-altmans-home-2026-04-13/}
}

@misc{marcus_substack,
  author       = {Marcus, Gary},
  title        = {Marcus on {AI}},
  howpublished = {Substack newsletter.
                  \url{https://garymarcus.substack.com/}},
  year         = {2022}
}

@misc{marcus2026scam,
  author       = {Marcus, Gary},
  title        = {Turns out Generative {AI} was a scam},
  howpublished = {Marcus on AI (Substack)},
  year         = {2026},
  month        = {February},
  day          = {23},
  url          = {https://garymarcus.substack.com/p/turns-out-generative-ai-was-a-scam}
}

@misc{marcus2024collapse,
  author       = {Marcus, Gary},
  title        = {Why the collapse of the Generative {AI} bubble may be imminent},
  howpublished = {Marcus on AI (Substack)},
  year         = {2024},
  month        = {August},
  day          = {3},
  url          = {https://garymarcus.substack.com/p/why-the-collapse-of-the-generative-ai-bubble-may-be-imminent}
}

@article{hoes2025distraction,
  author  = {Hoes, Emma and Gilardi, Fabrizio},
  title   = {Existential Risk Narratives about {AI} Do Not Distract
             from Its Immediate Harms},
  journal = {Proceedings of the National Academy of Sciences},
  volume  = {122},
  number  = {16},
  pages   = {e2419055122},
  year    = {2025},
  month   = apr,
  doi     = {10.1073/pnas.2419055122}
}

@article{hanna2023aiharm,
  author  = {Hanna, Alex and Bender, Emily M.},
  title   = {{AI} Causes Real Harm. Let's Focus on That over the
             End-of-Humanity Hype},
  journal = {Scientific American},
  volume  = {328},
  pages   = {69},
  year    = {2023}
}

@misc{hendrycks2025superintelligencestrategyexpertversion,
      title={Superintelligence Strategy: Expert Version}, 
      author={Dan Hendrycks and Eric Schmidt and Alexandr Wang},
      year={2025},
      eprint={2503.05628},
      archivePrefix={arXiv},
      primaryClass={cs.CY},
      url={https://arxiv.org/abs/2503.05628}, 
}

@book{suleyman2023wave,
  author    = {Suleyman, Mustafa and Bhaskar, Michael},
  title     = {The Coming Wave: Technology, Power, and the
               Twenty-First Century's Greatest Dilemma},
  year      = {2023},
  publisher = {Crown},
  address   = {New York},
  isbn      = {9780593593950}
}

@misc{carlini2026mythos,
  author       = {Carlini, Nicholas and Cheng, Newton and Lucas, Keane
                  and Moore, Michael and Nasr, Milad and Prabhushankar, Vinay
                  and Xiao, Winnie and Angulu, Hakeem and {Ben Asher}, Evyatar
                  and Bow, Jackie and Bradwell, Keir and Buchanan, Ben
                  and Forsythe, David and Freeman, Daniel and Gaynor, Alex
                  and Ge, Xinyang and Graham, Logan and Guru, Kyla
                  and Lakhani, Hasnain and McNiece, Matt and Mehrara, Mojtaba
                  and Nichol, Renee and Pirzada, Adnan and Porter, Sophia
                  and Terzis, Andreas and Troy, Kevin},
  title        = {Assessing {Claude Mythos Preview}'s Cybersecurity Capabilities},
  year         = {2026},
  month        = apr,
  day          = {7},
  howpublished = {Anthropic Red Team Blog.
                  \url{https://red.anthropic.com/2026/mythos-preview/}}
}

@misc{gtig2026may,
  author       = {{Google Threat Intelligence Group}},
  title        = {Adversaries Leverage {AI} for Vulnerability Exploitation,
                  Augmented Operations, and Initial Access},
  year         = {2026},
  month        = may,
  day          = {12},
  howpublished = {Google Cloud Blog.
                  \url{https://cloud.google.com/blog/topics/threat-intelligence/ai-vulnerability-exploitation-initial-access}}
}

@misc{kumar2025quantifyingcbrnriskfrontier,
      title={Quantifying CBRN Risk in Frontier Models}, 
      author={Divyanshu Kumar and Nitin Aravind Birur and Tanay Baswa and Sahil Agarwal and Prashanth Harshangi},
      year={2025},
      eprint={2510.21133},
      archivePrefix={arXiv},
      primaryClass={cs.CR},
      url={https://arxiv.org/abs/2510.21133}, 
}

@techreport{microsoft2024threats,
  author      = {{Microsoft Threat Intelligence} and {OpenAI}},
  title       = {Staying Ahead of Threat Actors in the Age of {AI}},
  institution = {Microsoft},
  year        = {2024},
  month       = feb,
  day         = {14},
  url         = {https://www.microsoft.com/en-us/security/blog/2024/02/14/staying-ahead-of-threat-actors-in-the-age-of-ai/}
}

@misc{long2024takingaiwelfareseriously,
      title={Taking AI Welfare Seriously}, 
      author={Robert Long and Jeff Sebo and Patrick Butlin and Kathleen Finlinson and Kyle Fish and Jacqueline Harding and Jacob Pfau and Toni Sims and Jonathan Birch and David Chalmers},
      year={2024},
      eprint={2411.00986},
      archivePrefix={arXiv},
      primaryClass={cs.CY},
      url={https://arxiv.org/abs/2411.00986}, 
}

@misc{anthropic2025welfare,
  author       = {{Anthropic}},
  title        = {Exploring Model Welfare},
  year         = {2025},
  month        = apr,
  day          = {24},
  howpublished = {\url{https://www.anthropic.com/research/exploring-model-welfare}}
}

@misc{perrigo2026teach,
  author       = {Perrigo, Billy},
  title        = {Can You Teach an {AI} to Be Good? {Anthropic} Thinks So},
  year         = {2026},
  month        = jan,
  day          = {21},
  howpublished = {\textit{Time}.
                  \url{https://time.com/7354738/claude-constitution-ai-alignment/}}
}

@misc{askell2026constitution,
  author       = {Askell, Amanda and Carlsmith, Joe and Olah, Chris
                  and Kaplan, Jared and Karnofsky, Holden
                  and {many other contributors}},
  title        = {Claude's Constitution},
  year         = {2026},
  month        = jan,
  day          = {21},
  howpublished = {Anthropic. \url{https://www.anthropic.com/constitution}}
}

@misc{hinton2025lbc,
  author       = {Hinton, Geoffrey},
  title        = {Interview on \textit{Tonight with Andrew Marr}},
  howpublished = {LBC. \url{https://www.lbc.co.uk/article/ai-consciousness-geoffrey-hinton-5HjdRXD_2/}},
  year         = {2025},
  month        = jan,
  day          = {30}
}

@misc{sutskever2022conscious,
  author       = {Sutskever, Ilya},
  title        = {``it may be that today's large neural networks
                  are slightly conscious''},
  howpublished = {Post on {X} (formerly Twitter).
                  \url{https://x.com/ilyasut/status/1491554478243258368}},
  year         = {2022},
  month        = feb,
  day          = {9}
}

@article{aguera2022llms,
  author    = {{Agüera y Arcas}, Blaise},
  title     = {Do Large Language Models Understand Us?},
  journal   = {Daedalus},
  volume    = {151},
  number    = {2},
  pages     = {183--197},
  year      = {2022},
  publisher = {MIT Press},
  doi       = {10.1162/daed_a_01909}
}

@misc{rabanser2026scienceaiagentreliability,
      title={Towards a Science of AI Agent Reliability}, 
      author={Stephan Rabanser and Sayash Kapoor and Peter Kirgis and Kangheng Liu and Saiteja Utpala and Arvind Narayanan},
      year={2026},
      eprint={2602.16666},
      archivePrefix={arXiv},
      primaryClass={cs.AI},
      url={https://arxiv.org/abs/2602.16666}, 
}

@misc{kaplan2020scalinglawsneurallanguage,
      title={Scaling Laws for Neural Language Models}, 
      author={Jared Kaplan and Sam McCandlish and Tom Henighan and Tom B. Brown and Benjamin Chess and Rewon Child and Scott Gray and Alec Radford and Jeffrey Wu and Dario Amodei},
      year={2020},
      eprint={2001.08361},
      archivePrefix={arXiv},
      primaryClass={cs.LG},
      url={https://arxiv.org/abs/2001.08361}, 
}

@article{lu2026automation,
  author    = {Lu, Chris and Lu, Cong and Lange, Robert Tjarko
               and Yamada, Yutaro and Hu, Shengran and Foerster, Jakob
               and Ha, David and Clune, Jeff},
  title     = {Towards end-to-end automation of {AI} research},
  journal   = {Nature},
  volume    = {651},
  number    = {8107},
  pages     = {914--919},
  year      = {2026},
  publisher = {Nature Publishing Group},
  doi       = {10.1038/s41586-026-10265-5}
}

@article{acemoglu2025macroeconomics,
  author    = {Acemoglu, Daron},
  title     = {The simple macroeconomics of {AI}},
  journal   = {Economic Policy},
  volume    = {40},
  number    = {121},
  pages     = {13--58},
  year      = {2025},
  publisher = {Oxford University Press},
  doi       = {10.1093/epolic/eiae042}
}

@article{brynjolfsson2021jcurve,
  author    = {Brynjolfsson, Erik and Rock, Daniel and Syverson, Chad},
  title     = {The Productivity {J}-Curve: How Intangibles Complement
               General Purpose Technologies},
  journal   = {American Economic Journal: Macroeconomics},
  volume    = {13},
  number    = {1},
  pages     = {333--372},
  year      = {2021},
  publisher = {American Economic Association},
  doi       = {10.1257/mac.20180386}
}

@misc{darioHype,
    url={https://darioamodei.com/essay/machines-of-loving-grace},
    author={Dario Amodei},
    title={Machines of Loving Grace},
    year={2024},
    month={October}
}

@misc{gentleSingularity,
    url={https://blog.samaltman.com/the-gentle-singularity},
    author={Sam Altman},
    title={The Gentle Singularity},
    year={2025},
    month={June}
}

@misc{andreessen2023techno,
  author       = {Andreessen, Marc},
  title        = {The Techno-Optimist Manifesto},
  year         = {2023},
  month        = oct,
  day          = {16},
  howpublished = {Andreessen Horowitz. \url{https://a16z.com/the-techno-optimist-manifesto/}}
}

@misc{rogelberg2026musk,
  author       = {Rogelberg, Sasha},
  title        = {Elon {Musk} Says That in 10 to 20 Years, Work Will Be
                  Optional and Money Will Be Irrelevant Thanks to {AI}
                  and Robotics},
  year         = {2026},
  month        = jan,
  day          = {19},
  howpublished = {\textit{Fortune}.
                  \url{https://fortune.com/2026/01/19/when-does-elon-musk-say-work-will-be-optional-and-money-will-be-irrelevant-ai-robotics/}}
}

@book{kurzweil2005singularity,
  author    = {Kurzweil, Ray},
  title     = {The Singularity {Is} Near: When Humans Transcend Biology},
  year      = {2005},
  publisher = {Viking},
  address   = {New York},
  isbn      = {9780670033843}
}

@misc{openai2018charter,
  author       = {{OpenAI}},
  title        = {{OpenAI} Charter},
  year         = {2018},
  month        = apr,
  howpublished = {\url{https://openai.com/charter/}}
}

@misc{anthropic2021company,
  author       = {{Anthropic}},
  title        = {Company},
  howpublished = {\url{https://www.anthropic.com/company}},
  note         = {Accessed May 21, 2026}
}

@misc{khosla2024roadmap,
  author       = {Khosla, Vinod},
  title        = {A Roadmap to {AI} Utopia},
  year         = {2024},
  month        = nov,
  day          = {11},
  howpublished = {\textit{Time}. \url{https://time.com/7174892/a-roadmap-to-ai-utopia/}}
}

@misc{intelligenceAge,
  author       = {Altman, Sam},
  title        = {The Intelligence Age},
  year         = {2024},
  month        = sep,
  day          = {23},
  howpublished = {\url{https://ia.samaltman.com}}
}

@inproceedings{amdahl1967validity,
  author    = {Amdahl, Gene M.},
  title     = {Validity of the Single Processor Approach to Achieving
               Large Scale Computing Capabilities},
  booktitle = {Proceedings of the {AFIPS} Spring Joint Computer Conference},
  series    = {{AFIPS} '67 (Spring)},
  pages     = {483--485},
  year      = {1967},
  publisher = {AFIPS Press},
  address   = {Reston, VA}
}

@article{drugDiscovery,
    author = {Lenarczyk, Gabriela and Minssen, Timo and Price, Nicholson and Rai, Arti},
    title = {The future of AI regulation in drug development: a comparative analysis},
    journal = {Journal of Law and the Biosciences},
    volume = {12},
    number = {2},
    pages = {lsaf028},
    year = {2025},
    month = {07},
    issn = {2053-9711},
    doi = {10.1093/jlb/lsaf028},
    url = {https://doi.org/10.1093/jlb/lsaf028},
    eprint = {https://academic.oup.com/jlb/article-pdf/12/2/lsaf028/65244130/lsaf028.pdf},
}

@misc{thais2024misrepresentedtechnologicalsolutionsimagined,
      title={Misrepresented Technological Solutions in Imagined Futures: The Origins and Dangers of AI Hype in the Research Community}, 
      author={Savannah Thais},
      year={2024},
      eprint={2408.15244},
      archivePrefix={arXiv},
      primaryClass={cs.CY},
      url={https://arxiv.org/abs/2408.15244}, 
}

@book{crawford2021atlas,
  author    = {Crawford, Kate},
  title     = {Atlas of {AI}: Power, Politics, and the Planetary Costs
               of Artificial Intelligence},
  year      = {2021},
  publisher = {Yale University Press},
  address   = {New Haven, CT},
  isbn      = {9780300209570}
}

@book{narayanan2024snakeoil,
  author    = {Narayanan, Arvind and Kapoor, Sayash},
  title     = {{AI} Snake Oil: What Artificial Intelligence Can Do,
               What It Can't, and How to Tell the Difference},
  year      = {2024},
  publisher = {Princeton University Press},
  address   = {Princeton, NJ},
  isbn      = {9780691249131}
}

@techreport{ainowinstitute2018,
  author      = {Whittaker, Meredith and Crawford, Kate and Dobbe, Roel
                 and Fried, Genevieve and Kaziunas, Elizabeth and Mathur, Varoon
                 and {Myers West}, Sarah and Richardson, Rashida and Schultz, Jason
                 and Schwartz, Oscar},
  title       = {{AI} Now 2018 Report},
  institution = {{AI} Now Institute},
  address     = {New York},
  year        = {2018},
  month       = dec,
  day         = {6},
  url         = {https://ainowinstitute.org/AI_Now_2018_Report.html}
}

@techreport{briggs2023ai,
  author      = {Briggs, Joseph and Kodnani, Devesh},
  title       = {The Potentially Large Effects of Artificial Intelligence
                 on Economic Growth},
  institution = {Goldman Sachs Economics Research},
  year        = {2023},
  month       = mar,
  url         = {https://www.goldmansachs.com/intelligence/pages/generative-ai-could-raise-global-gdp-by-7-percent.html}
}

@misc{kosmyna2025brainchatgptaccumulationcognitive,
      title={Your Brain on ChatGPT: Accumulation of Cognitive Debt when Using an AI Assistant for Essay Writing Task}, 
      author={Nataliya Kosmyna and Eugene Hauptmann and Ye Tong Yuan and Jessica Situ and Xian-Hao Liao and Ashly Vivian Beresnitzky and Iris Braunstein and Pattie Maes},
      year={2025},
      eprint={2506.08872},
      archivePrefix={arXiv},
      primaryClass={cs.AI},
      url={https://arxiv.org/abs/2506.08872}, 
}

@article{gerlich2025ai,
  author    = {Gerlich, Michael},
  title     = {{AI} Tools in Society: Impacts on Cognitive Offloading
               and the Future of Critical Thinking},
  journal   = {Societies},
  volume    = {15},
  number    = {1},
  pages     = {6},
  year      = {2025},
  publisher = {MDPI},
  doi       = {10.3390/soc15010006}
}

@misc{fang2025aihumanbehaviorsshape,
      title={How AI and Human Behaviors Shape Psychosocial Effects of Extended Chatbot Use: A Longitudinal Randomized Controlled Study}, 
      author={Cathy Mengying Fang and Auren R. Liu and Valdemar Danry and Eunhae Lee and Samantha W. T. Chan and Pat Pataranutaporn and Pattie Maes and Jason Phang and Michael Lampe and Lama Ahmad and Sandhini Agarwal},
      year={2025},
      eprint={2503.17473},
      archivePrefix={arXiv},
      primaryClass={cs.HC},
      url={https://arxiv.org/abs/2503.17473}, 
}

@misc{phang2025investigatingaffectiveuseemotional,
      title={Investigating Affective Use and Emotional Well-being on ChatGPT}, 
      author={Jason Phang and Michael Lampe and Lama Ahmad and Sandhini Agarwal and Cathy Mengying Fang and Auren R. Liu and Valdemar Danry and Eunhae Lee and Samantha W. T. Chan and Pat Pataranutaporn and Pattie Maes},
      year={2025},
      eprint={2504.03888},
      archivePrefix={arXiv},
      primaryClass={cs.HC},
      url={https://arxiv.org/abs/2504.03888}, 
}

@article{li2025thirsty,
  author    = {Li, Pengfei and Yang, Jianyi and Islam, Mohammad A. and Ren, Shaolei},
  title     = {Making {AI} Less ``Thirsty'': Uncovering and Addressing
               the Secret Water Footprint of {AI} Models},
  journal   = {Communications of the {ACM}},
  year      = {2025},
  publisher = {ACM},
  doi       = {10.1145/3724499}
}

@techreport{iea2025electricity,
  author      = {{International Energy Agency}},
  title       = {Electricity 2025: Analysis and Forecast to 2027},
  institution = {IEA},
  address     = {Paris},
  year        = {2025},
  url         = {https://www.iea.org/reports/electricity-2025}
}

@article{gallegos2024bias,
  author    = {Gallegos, Isabel O. and Rossi, Ryan A. and Barrow, Joe
               and Tanjim, Md Mehrab and Kim, Sungchul and Dernoncourt, Franck
               and Yu, Tong and Zhang, Ruiyi and Ahmed, Nesreen K.},
  title     = {Bias and Fairness in Large Language Models: {A} Survey},
  journal   = {Computational Linguistics},
  volume    = {50},
  number    = {3},
  pages     = {1097--1179},
  year      = {2024},
  publisher = {MIT Press},
  doi       = {10.1162/coli_a_00524}
}

@inproceedings{genghosts,
author = {Morris, Meredith Ringel and Brubaker, Jed R.},
title = {Generative Ghosts: Anticipating Benefits and Risks of AI Afterlives},
year = {2025},
isbn = {9798400713941},
publisher = {Association for Computing Machinery},
address = {New York, NY, USA},
url = {https://doi.org/10.1145/3706598.3713758},
doi = {10.1145/3706598.3713758},
booktitle = {Proceedings of the 2025 CHI Conference on Human Factors in Computing Systems},
articleno = {536},
numpages = {14},
location = {
},
series = {CHI '25}
}

@article{lampost,
author = {Goodman, Steven M. and Buehler, Erin and Clary, Patrick and Coenen, Andy and Donsbach, Aaron and Horne, Tiffanie N. and Lahav, Michal and MacDonald, Robert and Michaels, Rain Breaw and Narayanan, Ajit and Pushkarna, Mahima and Riley, Joel and Santana, Alex and Shi, Lei and Sweeney, Rachel and Weaver, Phil and Yuan, Ann and Morris, Meredith Ringel},
title = {LaMPost: AI Writing Assistance for Adults with Dyslexia Using Large Language Models},
year = {2024},
issue_date = {September 2024},
publisher = {Association for Computing Machinery},
address = {New York, NY, USA},
volume = {67},
number = {9},
issn = {0001-0782},
url = {https://doi.org/10.1145/3626952},
doi = {10.1145/3626952},
journal = {Commun. ACM},
month = aug,
pages = {80–89},
numpages = {10}
}

@article{noy2023experimental,
  author    = {Noy, Shakked and Zhang, Whitney},
  title     = {Experimental evidence on the productivity effects of
               generative artificial intelligence},
  journal   = {Science},
  volume    = {381},
  number    = {6654},
  pages     = {187--192},
  year      = {2023},
  publisher = {American Association for the Advancement of Science},
  doi       = {10.1126/science.adh2586}
}

@article{brynjolfsson2025generative,
  author    = {Brynjolfsson, Erik and Li, Danielle and Raymond, Lindsey},
  title     = {Generative {AI} at Work},
  journal   = {The Quarterly Journal of Economics},
  volume    = {140},
  number    = {2},
  pages     = {889--942},
  year      = {2025},
  publisher = {Oxford University Press},
  doi       = {10.1093/qje/qjae044}
}

@article{paglialunga2025effectiveness,
  author    = {Paglialunga, Andrea and Melogno, Sergio},
  title     = {The Effectiveness of Artificial Intelligence-Based
               Interventions for Students with Learning Disabilities:
               {A} Systematic Review},
  journal   = {Brain Sciences},
  volume    = {15},
  number    = {8},
  pages     = {806},
  year      = {2025},
  publisher = {MDPI},
  doi       = {10.3390/brainsci15080806}
}

@article{defreitas2025ai,
  author    = {De Freitas, Julian and U{\u{g}}uralp, Ahmet Kaan
               and U{\u{g}}uralp, Zeliha and Puntoni, Stefano},
  title     = {{AI} Companions Reduce Loneliness},
  journal   = {Journal of Consumer Research},
  year      = {2025},
  publisher = {Oxford University Press},
  doi       = {10.1093/jcr/ucaf040}
}

@article{bai2023ai,
  author    = {Bai, Fang and Li, Shiliang and Li, Honglin},
  title     = {{AI} enhances drug discovery and development},
  journal   = {National Science Review},
  volume    = {11},
  number    = {3},
  pages     = {nwad303},
  year      = {2023},
  publisher = {Oxford University Press},
  doi       = {10.1093/nsr/nwad303}
}

@article{jumper2021alphafold,
  author    = {Jumper, John and Evans, Richard and Pritzel, Alexander
               and Green, Tim and Figurnov, Michael and Ronneberger, Olaf
               and Tunyasuvunakool, Kathryn and Bates, Russ
               and {\v{Z}}{\'{i}}dek, Augustin and Potapenko, Anna
               and Bridgland, Alex and Meyer, Clemens
               and Kohl, Simon A. A. and Ballard, Andrew J.
               and Cowie, Andrew and Romera-Paredes, Bernardino
               and Nikolov, Stanislav and Jain, Rishub and Adler, Jonas
               and Back, Trevor and Petersen, Stig and Reiman, David
               and Clancy, Ellen and Zielinski, Michal
               and Steinegger, Martin and Pacholska, Michalina
               and Berghammer, Tamas and Bodenstein, Sebastian
               and Silver, David and Vinyals, Oriol and Senior, Andrew W.
               and Kavukcuoglu, Koray and Kohli, Pushmeet
               and Hassabis, Demis},
  title     = {Highly accurate protein structure prediction with
               {AlphaFold}},
  journal   = {Nature},
  volume    = {596},
  number    = {7873},
  pages     = {583--589},
  year      = {2021},
  publisher = {Springer Nature},
  doi       = {10.1038/s41586-021-03819-2}
}

@inproceedings{whatIsLiteracy,
author = {Long, Duri and Magerko, Brian},
title = {What is AI Literacy? Competencies and Design Considerations},
year = {2020},
isbn = {9781450367080},
publisher = {Association for Computing Machinery},
address = {New York, NY, USA},
url = {https://doi.org/10.1145/3313831.3376727},
doi = {10.1145/3313831.3376727},
booktitle = {Proceedings of the 2020 CHI Conference on Human Factors in Computing Systems},
pages = {1–16},
numpages = {16},
location = {Honolulu, HI, USA},
series = {CHI '20}
}

@article{AIMagLiteracy,
    title={AI literacy as a core component of AI education},
    year={2025},
    month={July},
    journal={AI Magazine},
    author = {Sri Yash Tadimalla and Mary Lou Maher},
    doi={https://doi.org/10.1002/aaai.70007},
    volume={46},
    issue={2}
}

@misc{OECD2025,
    author = {OECD},
    title = {Bridging the AI skills gap: Is training keeping up?},
    doi={https://doi.org/10.1787/66d0702e-en},
    publisher={OECD Publishing, Paris},
    year = {2025}
}

@article{LAUPICHLER2022100101,
title = {Artificial intelligence literacy in higher and adult education: A scoping literature review},
journal = {Computers and Education: Artificial Intelligence},
volume = {3},
pages = {100101},
year = {2022},
issn = {2666-920X},
doi = {https://doi.org/10.1016/j.caeai.2022.100101},
url = {https://www.sciencedirect.com/science/article/pii/S2666920X2200056X},
author = {Matthias Carl Laupichler and Alexandra Aster and Jana Schirch and Tobias Raupach}
}

@techreport{maslej2025aiindex,
  author      = {Maslej, Nestor and Fattorini, Loredana and Perrault, Raymond
                 and Gil, Yolanda and Parli, Vanessa and Kariuki, Njenga
                 and Capstick, Emily and Reuel, Anka and Brynjolfsson, Erik
                 and Etchemendy, John and Ligett, Katrina and Lyons, Terah
                 and Manyika, James and Niebles, Juan Carlos and Shoham, Yoav
                 and Wald, Russell and Walsh, Toby and Hamrah, Armin
                 and Santarlasci, Lapo and {Betts Lotufo}, Julia
                 and Rome, Alexandra and Shi, Andrew and Oak, Sukrut},
  title       = {Artificial Intelligence Index Report 2025},
  institution = {AI Index Steering Committee, Institute for Human-Centered {AI},
                 Stanford University},
  address     = {Stanford, CA},
  year        = {2025},
  url         = {https://hai.stanford.edu/ai-index/2025-ai-index-report}
}

@article{EUsovereign,
    author = {Daniel Mügge},
    title = {EU AI sovereignty: for whom, to what end, and to whose benefit?},
    journal = {Journal of European Public Policy},
    volume = {31},
    number = {8},
    pages = {2200--2225},
    year = {2024},
    publisher = {Routledge},
    doi = {10.1080/13501763.2024.2318475}

}

@techreport{pew2026datacenters,
   author      = {Gramlich, John and Kennedy, Brian and McClain, Colleen
                 and Stocking, Galen},
  title       = {How {Americans} View Data Centers' Impact in Key Areas,
                 from the Environment to Jobs},
  institution = {Pew Research Center},
  year        = {2026},
  month       = mar,
  url         = {https://www.pewresearch.org/short-reads/2026/03/12/how-americans-view-data-centers-impact-in-key-areas-from-the-environment-to-jobs/}
}

@techreport{pew2025public,
  author      = {McClain, Colleen and Kennedy, Brian and Gottfried, Jeffrey
                 and Anderson, Monica and Pasquini, Giancarlo},
  title       = {How the {US} Public and {AI} Experts View Artificial Intelligence},
  institution = {Pew Research Center},
  year        = {2025},
  month       = apr,
  url         = {https://www.pewresearch.org/internet/2025/04/03/how-the-us-public-and-ai-experts-view-artificial-intelligence/}
}

@misc{axios2026harrispoll,
  author       = {{Axios} and {The Harris Poll}},
  title        = {{2026 Axios Harris Poll 100}: {GOP} Embraces {AI} Over Democrats},
  year         = {2026},
  month        = may,
  howpublished = {Axios},
  url          = {https://www.axios.com/2026/05/19/axios-harris-poll-100-ai-politics}
}

@misc{perceivedConsciousness,
      title={AI and Consciousness: Shifting Focus Towards Tractable Questions}, 
      author={Iulia-Maria Comsa},
      year={2026},
      eprint={2605.06965},
      archivePrefix={arXiv},
      primaryClass={cs.CY},
      url={https://arxiv.org/abs/2605.06965}, 
}

@book{beingYou,
    author = {Anil Seth},
    title = {Being You: A New Science of Consciousness},
    publisher = {Faber \& Faber},
    year = {2022}
}

@article{merrieJaggedness,
    author = {Morris Ringel Morris and Dan Altman and Haydn Belfield and Arthur Goemans and Hasan Iqbal and Ryan Burnell and Iason Gabriel and Samuel Albanie and Allan Dafoe},
    title = {Characterizing Jaggedness Aids Safety \& Usability},
    publisher = {Google DeepMind Technical Report},
    year = {2026},
    url = {https://cs.stanford.edu/~merrie/papers/jaggedness_preprint.pdf}
}

@misc{brooks7sins,
    author = {Rodney Brooks},
    title = {The Seven Deadly Sins of AI Predictions},
    journal = {MIT Technology Review},
    year = {2017},
    month = {October},
    url = {https://www.technologyreview.com/2017/10/06/241837/the-seven-deadly-sins-of-ai-predictions/}
}

@book{aiConBook,
    author = {Emily M. Bender and Alex Hanna},
    title = {The AI Con: How to Fight Big Tech's Hype and Create the Future We Want},
    publisher = {Harper},
    year = {2025},
    month = {May}
}

@inproceedings{parrots,
author = {Bender, Emily M. and Gebru, Timnit and McMillan-Major, Angelina and Shmitchell, Shmargaret},
title = {On the Dangers of Stochastic Parrots: Can Language Models Be Too Big?},
year = {2021},
isbn = {9781450383097},
publisher = {Association for Computing Machinery},
address = {New York, NY, USA},
url = {https://doi.org/10.1145/3442188.3445922},
doi = {10.1145/3442188.3445922},
booktitle = {Proceedings of the 2021 ACM Conference on Fairness, Accountability, and Transparency},
pages = {610–623},
numpages = {14},
series = {FAccT '21}
}

@book{rebootingAI,
    author = {Gary Marcus and Ernest Davis},
    title = {Rebooting AI: Building Artificial Intelligence We Can Trust},
    publisher = {Pantheon},
    year = {2019},
    month = {September}
}

@misc{hinton2025cbsnews,
  author       = {Hinton, Geoffrey},
  title        = {``{Godfather} of {AI}'' {Geoffrey Hinton} Warns {AI}
                  Could Take Control from Humans},
  year         = {2025},
  howpublished = {CBS News.
                  \url{https://www.cbsnews.com/news/godfather-of-ai-geoffrey-hinton-ai-warning}}
}

@misc{aiSafetyPetition,
    url={https://aistatement.com/#open-letter},
    month={May},
    year={2023},
    author={{Center for AI Safety}},
    title={Statement on AI Risk}
}
\bibliographystyle{icml2026}

\newpage
\appendix
\onecolumn



\end{document}